\documentclass[10pt,twocolumn,letterpaper]{article}

\usepackage{cvpr}
\usepackage{times}
\usepackage{epsfig}
\usepackage{graphicx}
\usepackage{amsmath}
\usepackage{amssymb}
\usepackage{enumitem}
\usepackage{mathrsfs}
\usepackage{autobreak}
\usepackage{booktabs}
\usepackage{threeparttable}
\usepackage{multirow}
\usepackage[T1]{fontenc}
\usepackage{algorithm}

\usepackage[pagebackref=true,breaklinks=true,letterpaper=true,colorlinks,bookmarks=false]{hyperref}

\cvprfinalcopy % *** Uncomment this line for the final submission

\def\cvprPaperID{8159} % *** Enter the CVPR Paper ID here

\ifcvprfinal\fi
\begin{document}

%%%%%%%%% TITLE
\title{Learning geometry-image representation for 3D point cloud generation}

\author{Lei Wang, Yuchun Huang, Pengjie Tao, Yaolin Hou, Yuxuan Liu \\
	%\textsuperscript{1}Wuhan University\\
	Wuhan University\\
	{\tt\small \{wlei, hycwhu, pjtao, houyaolin, liuyx\}@whu.edu.cn }
	% For a paper whose authors are all at the same institution,
	% omit the following lines up until the closing ``}''.
	% Additional authors and addresses can be added with ``\and'',
	% just like the second author.
	% To save space, use either the email address or home page, not both
}

\maketitle
%\thispagestyle{empty}

%%%%%%%%% ABSTRACT
\begin{abstract}
   We study the problem of generating point clouds of 3D objects. Instead of discretizing the object into 3D voxels with huge computational cost and resolution limitations, we propose a novel geometry image based generator (GIG) to convert the 3D point cloud generation problem to a 2D geometry image generation problem. Since the geometry image is a completely regular 2D array that contains the surface points of the 3D object, it leverages both the regularity of the 2D array and the geodesic neighborhood of the 3D surface. Thus, one significant benefit of our GIG is that it allows us to directly generate the 3D point clouds using efficient 2D image generation networks. Experiments on both rigid and non-rigid 3D object datasets have demonstrated the promising performance of our method to not only create plausible and novel 3D objects, but also learn a probabilistic latent space that well supports the shape editing like interpolation and arithmetic.
\end{abstract}

%%%%%%%%% BODY TEXT
\section{Introduction}

3D object generation is the problem of creating plausible and varied 3D objects under certain prior distributions and/or conditions. It is an important yet challenging task in computer vision and graphics, and has many real-world applications, such as virtual/augmented reality, dynamic interaction, and so on. Recent attempts at using deep learning for 3D object generation mainly rely on the volumetric grids~\cite{3D_GAN, wang2018global} and multi-view images~\cite{choy20163d_r2n2, lin2018learning}, but they are resolution-limited or occlusion-sensitive. In this work, we focus on generating 3D objects represented as point clouds.

However, deep learning architectures for 3D point cloud generation still face enormous challenges~\cite{l_GAN, pc_GAN}. Because of the irregular data structure of the 3D point cloud~\cite{PointNet}, convolution and deconvolution on grid domains (e.g., images and videos) are hard to generalize to point clouds for feature encoding and reconstruction. In addition, the dimensional complexity of point clouds in 3D space also brings challenges for the generation task. 

\begin{figure}[t]
	\setlength{\abovecaptionskip}{1cm} 
	\setlength{\belowcaptionskip}{-1.0cm} 
	\begin{center}
		\includegraphics[width=1\linewidth]{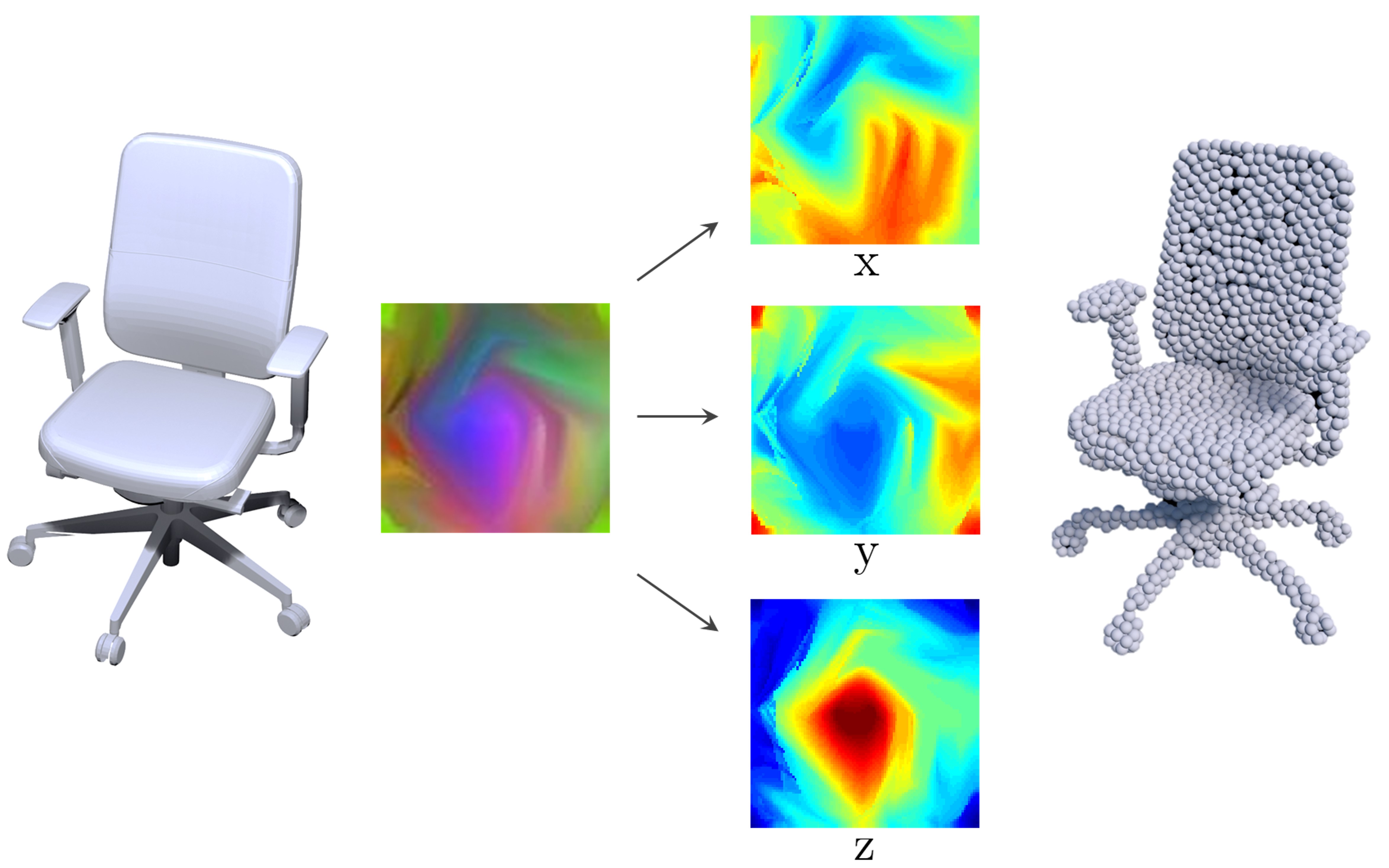}
	\end{center}
	\vspace{-.1cm}
	\caption{2D geometry image of a chair. Each pixel corresponds to a resampled surface point of the chair, containing rich information like ${\rm x}$, ${\rm y}$, ${\rm z}$ coordinates and even colors. Thus, the geometry image actually corresponds to a point cloud of the chair, leveraging both the regularity of the 2D array and the geodesic neighborhood of the 3D object surface.}
	\label{fig1}
\end{figure}

In fact, any 3D surface of the object can be resampled onto a regular plane called geometry image~\cite{gu2002geometry} (Figure~\ref{fig1}). It is a simple yet completely regular 2D array, containing the coordinate values of the 3D object's surface points, i.e., the point cloud. Thus the geometry image leverages both the regularity of the 2D array and the geodesic neighborhood of the 3D surface~\cite{cross_atlas}, convolution and deconvolution on 2D images can be easily extended onto it. Therefore, instead of directly generating the 3D point cloud, the key idea of this work is to instead generate its corresponding 2D geometry image with a novel GIG. 

By converting the 3D point cloud generation problem onto regular 2D geometry image, our GIG yields enormous benefits: 1) The geometry image inherits the geodesic neighborhood of the 3D object surface, convolution on the geometry image is directly equivalent to feature learning on the manifold surface of the 3D object; 2) The pyramid of geometry image is analogous to the sampling of 3D point cloud, making multi-resolution learning on point cloud easily achievable; 3) The architecture of our GIG can be constructed by referring to the classical 2D generation networks, which are more efficient compared to the direct deep learning on irregular 3D point cloud.

Correspondingly, our discriminator can be constructed by imitating the previous point cloud deep learning works like PointNet~\cite{PointNet}. However, because of the asymmetrical architectures of PointNet and our GIG, the classical generative adversarial network (GAN)~\cite{GAN} can hardly balance their training progresses. To this end, an adversarial VAE combining the adversarial learning of GAN with variational autoencoders (VAE)~\cite{VAE} is further introduced, which leverages both training stability and sample fidelity. 

To verify the effectiveness of our method, we evaluate it for both rigid and non-rigid 3D object generation on public datasets ShapeNet~\cite{ShapeNet} and D-FAUST~\cite{D_FAUST}. Experiments show that the proposed GIG is highly effective at capturing the geometric structure of the 3D object for plausible and novel 3D point cloud generation. In addition, like 2D images, the learned latent representations of the 3D point clouds also well support the shape editing like interpolation and arithmetic operations (e.g., addition and subtraction).

Overall, our contributions are threefold:
\begin{itemize}
	\vspace{-.2cm}
	\item We propose a novel geometry image based generator to convert the 3D point cloud generation problem onto regular 2D grid, so that the complex 3D point cloud can be generated by referring to simple 2D generation networks;
	\vspace{-.2cm}
	\item We introduce a novel adversarial VAE to optimize the proposed GIG by combining adversarial learning with VAE~\cite{VAE};
	\vspace{-.2cm}
	\item We train an end-to-end 3D point cloud generation model with the proposed GIG and demonstrate its effectiveness with large-scale experiments on both rigid and non-rigid 3D object datasets.
\end{itemize}

%-------------------------------------------------------------------------
\section{Related Works}

This section will discuss the related prior works in three main aspects: deep learning for point cloud analysis, 3D generative networks, and resampling 3D surfaces to planar domain.

\textbf{Deep learning for point cloud analysis.}
Recently, researchers have attempted to generalize convolutional neural networks (CNNs) to unorganized 3D point clouds including 1) voxelization-based methods~\cite{pointgrid,voxnet,O_CNN, 3d_ShapeNets,OctNet} which discretize the 3D space into regular voxels so that the 3D convolutional network can be similarly applied. However, volumetric representation inevitably leads to the discarding of small details as well as high memory usage and computational cost. 2) Multi-view projection~\cite{su2015multi,multiview_mesh,project_CNN,RotationNet} renders the images from multiple views of the 3D object. This allows the standard 2D CNNs to be directly applied, but it is still difficult to determine the distribution and number of the views to cover the entire 3D object while avoiding mutual occlusions. 3) Graph convolution~\cite{geometric_DL,KCNet,dynamic_filter,GACNet,AnisotropicCNN,geometric_CNN} generalizes the standard convolution to graph-structured data and can be directly applied on mesh. For 3D point cloud, it should be first organized as a graph according to their spatial neighbors. 4) Set-based learning~\cite{PointNet,PointNet++,deep_sets,A_CNN} is a recent break-through which can directly process the unorganized point clouds. It allows researchers to construct effective and simple point cloud feature learning networks by first employing multi-layer perceptron (MLP) on each point and then aggregating them together as a global feature. However, the set-based methods neglect the local geometric structure of the 3D point cloud for fine-grained feature representation.

\begin{figure*}
	\setlength{\abovecaptionskip}{3pt} 
	\setlength{\belowcaptionskip}{-3pt} 
	\begin{center}
		\includegraphics[width=1.0\linewidth]{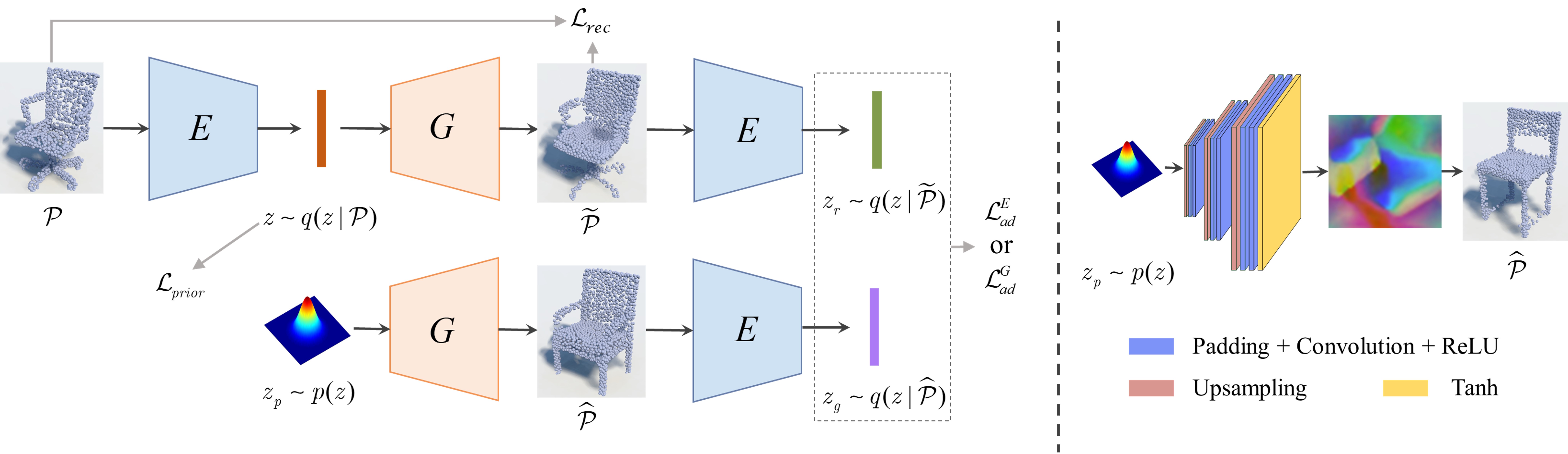}
	\end{center}
	\vspace{-0.2cm}
	\caption{\textbf{Left}: Framework of the proposed 3D point cloud generation model. It consists of an encoder \textit{E} and a generator \textit{G} as VAE~\cite{VAE}. However, in addition to minimizing the point-wise distance between the reconstructed and real point clouds, we further encourage them to have the same distribution in the latent space, which can be seen as a feature level constraint to ensure the fidelity of the generated point clouds. 
	\textbf{Right}: Architecture of the proposed GIG. By converting the 3D point cloud generation problem to the regular 2D geometry image, the proposed GIG can be simply constructed by referring to efficient 2D generation networks. It accepts a sample from the prior distribution and maps it to a geometry image with three channels, directly corresponding to the 3D coordinates of point cloud.  }
	\label{fig2}
\end{figure*}

\textbf{3D generative networks.}
Deep generative models~\cite{RelGAN,Stackgan,DCGAN,Introvae} based on GAN~\cite{GAN} and VAE~\cite{VAE} have shown compelling progress in image and text generation tasks. They aim to learn a generator that can create plausible images from given distributions. Recently, with the introduction of large public 3D model datasets like ShapeNet~\cite{ShapeNet}, there have been some inspiring attempts to generate 3D objects with similar techniques~\cite{3D_GAN,wang2018global,lin2018learning,l_GAN,3DAutoencoder,VAE_mesh}. The most intuitive way is to discretize the 3D object into regular voxel grids~\cite{3D_GAN,wang2018global,paired_3DGAN,dai2017shape,OctreeGAN} or project it into 2D images from multiple views~\cite{choy20163d_r2n2,lin2018learning,multiview_depth,Single2Multiview}, so that the image generation networks can be directly extended. However, the volumetric representation inevitably leads to (high) computational consumption and resolution limitation, while the multi-view projection is sensitive to the occlusions. 

For the representation of 3D point clouds, deep generative networks are still under explored and poses challenges~\cite{l_GAN, pc_GAN}. The generation model usually consists of two parts: a discriminator/encoder and a generator/decoder. Though previous deep learning works for point cloud analysis~\cite{PointNet,PointNet++,KCNet} have provided rich design references for the encoder, the 3D point cloud generator, which in turn needs to recover the details of the 3D objects, still faces challenges. 
Existing 3D point cloud generation models mainly use the fully-connected network (FCN) as the generator/decoder~\cite{l_GAN,pc_GAN,3DAutoencoder}, whose outputs are a series of 3D point coordinates of the point cloud. However, FCN requires an enormous amount of parameters when increasing the number of points and neglects the local geometric structure of the 3D point cloud. Recently, FoldingNet~\cite{FoldingNet} designed a decoder that learns a mapping from a 2D grid to the 3D point cloud under the guidance of a global feature vector. AtlasNet~\cite{AtlasNet} is also aimed at generating the point cloud as a manifold surface through a series of 2D grids. TopNet~\cite{TopNet} proposed a tree-structured decoder that generates the point cloud as the leaves of a binary tree, where each node is split from its father node. However, all these decoders are based on the point-wise MLP that neglects the local geometric structure of the point cloud, which is important for fine-grained 3D object generation. 

\textbf{Resampling 3D surfaces to planar domain.}
To resample a given 3D surface onto a 2D plane, traditional processes involve partitioning the surface into charts, individually parametrizing them into the planar domain, and then packing them together as a texture atlas~\cite{spherical_geometryimage}. However, the presence of visible seams on the reconstructed surface is always inevitable. In addition, because of the isolation of each atlas, their neighborhoods are actually taken apart from each other~\cite{cross_atlas}, which brings difficulties in applying the standard convolutional networks. 

Geometry images~\cite{gu2002geometry,DL_geometryimage} aim to resample the 3D surface onto a completely regular grid. It is a simple 2D array that stores rich information of the resampled 3D surface points like ${\rm x}$, ${\rm y}$, ${\rm z}$ coordinate values, colors and even normals. The resampling process involves cutting the 3D surface onto a plane and then mapping its boundary to a square. Therefore, the geometry image inherits the geodesic neighborhood of the 3D surface on a regular 2D grid, which is natural for feature learning with standard CNNs. 

The proposed GIG is inspired by the geometry-image representation of 3D object surface. However, instead of creating geometry images with traditional complex resampling algorithms~\cite{gu2002geometry,DL_geometryimage}, we implement this process with an end-to-end learning network in this paper. Thus, we actually construct a more generic 2D grid representation of the 3D object surface, and the traditional geometry image~\cite{gu2002geometry,DL_geometryimage} can be regarded as one particular solution of our generation model.

%-------------------------------------------------------------------------
\section{Method}
\label{method}

We propose a novel geometry image based generator (GIG) for 3D point cloud generation. By converting the 3D point cloud generation problem onto regular 2D  grid, the architecture of our GIG can be as simple as the classical 2D image generation networks (Section~\ref{network_architecture}). Afterwards, the proposed GIG is optimized under the criterion of the adversarial VAE (ad-VAE) which combines the adversarial learning with VAE (Section~\ref{ad-VAE}). 

Denote the given point cloud of an object as $\mathcal{P}=\{x_1,x_2,...,x_n\} \in \mathbb{R}^3$, the proposed 3D point cloud generation model is composed of two parts: the encoder $\textit{E}$ maps the point cloud $\mathcal{P}$ to the latent representation $z$ while the generator $\textit{G}$ runs the reverse process:
\begin{equation}
	z \sim \textit{E}(\mathcal{P}) = q(z|\mathcal{P}) ,\quad  \mathcal{\tilde{P}} \sim \textit{G}(z) = p(\mathcal{P}|z)
\end{equation}
Typically, the latent representation $z$ is regularized to a prior distribution $p(z)$, so that the generator $\textit{G}$ can map a given sample $z_p \sim p(z)$ to a plausible point cloud of a 3D object (Figure~\ref{fig2}). In the following sections, we start by introducing the architectures of our encoder and generator (Section~\ref{network_architecture}), and then their optimization details (Section~\ref{ad-VAE}). 

\subsection{Architectures of the 3D Generation Model}
\label{network_architecture}

\textbf{Structure guided encoder} \textit{E}.
Recent point cloud analysis networks~\cite{PointNet,KCNet,PointNet++} have provided rich design references for our encoder $\textit{E}$. Especially, PointNet~\cite{PointNet} provides an effective and simple architecture to directly learn on point sets by first computing the individual point features using per-point MLP and then aggregating them as a global presentation. However, the missing of local geometric structure in PointNet makes it unsuitable for fine-grained point cloud representation. To this end, we first capture the local geometric structures for each point using a set of learnable kernel points $\kappa=\{\kappa_l \}_{l=1}^L$ inspired by KCNet~\cite{KCNet}, and then concatenate them with the point coordinates as the input of PointNet~\cite{PointNet} to obtain the latent representation.

For each point $x_i\in \mathcal{P}$, we use the Gaussian kernel to measure the similarity between the point-set kernel $\kappa_l$ and its neighbor set as 
\begin{equation}
	K_\sigma(\kappa_l, x_i) = {\rm exp} \left( - \frac{d_{CD} \left(\kappa_l, \mathcal{N}_i \right) }{ 2 \sigma^2 } \right)
\end{equation}
where $\mathcal{N}_i$ is the neighbor points of $x_i$, and 
\begin{equation} \begin{split}
	d_{CD} (\kappa_l, \mathcal{N}_i) = %\frac{1}{2} (
	&\frac{1}{2|\mathcal{N}_i|} \sum_{x\in \mathcal{N}_i} \min \limits_{y\in \kappa_l} \|x-y\|^2 \\
	&+  \frac{1}{2|\kappa_l|} \sum_{y\in \kappa_l} \min \limits_{x\in \mathcal{N}_i} \|x-y\|^2 %)
\label{ChamferDistance}
\end{split} \end{equation}
is the Chamfer distance between $\kappa_l$ and $\mathcal{N}_i$, which is more effective at measuring the similarity between two point sets than KCNet~\cite{KCNet}. In addition, to adapt to the generation task, the output of PointNet is modified to the means and standard deviations as VAE~\cite{VAE}, and the latent representation is then obtained using reparameterization~\cite{VAE}.

\textbf{Geometry image based generator} \textit{G}.
Instead of directly generating the 3D point cloud, the proposed GIG aims to generate its 2D geometry-image representation by referring to the efficient 2D image generation networks. 

However, can we directly apply the classical 2D generation networks~\cite{DCGAN,Introvae} to the problem of generating geometry images? In the following, we discuss this from three main aspects of the 2D image generation networks including: convolution, padding and upsampling (with interpolation or deconvolution).

\begin{figure}
	\setlength{\abovecaptionskip}{1cm} 
	\setlength{\belowcaptionskip}{-1.0cm} 
	\begin{center}
		\includegraphics[width=1.0\linewidth]{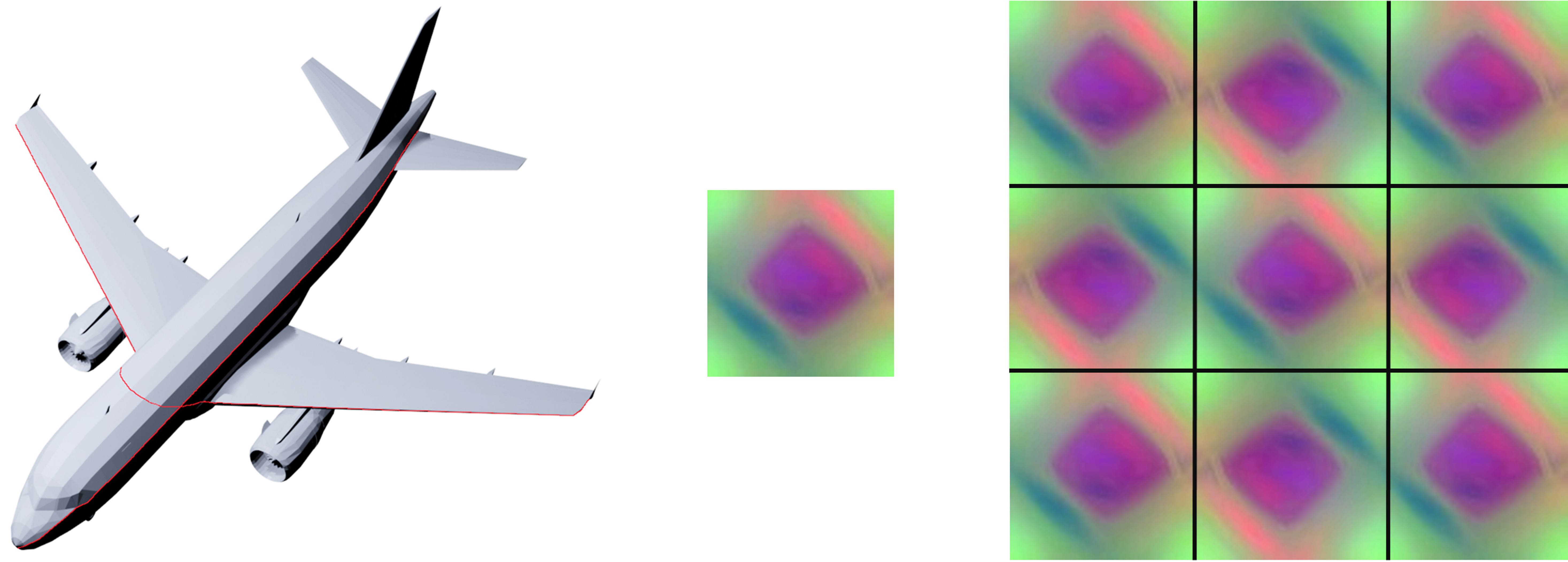}
	\end{center}
	\vspace{-0.2cm}
	\caption{Padding structure of the geometry image. The center image in the grid is the original geometry image. Its rotated images can be seamlessly connected to it, and this boundary property can be well preserved under the operation of convolution. }
	\label{padding}
\end{figure}

\begin{itemize}
	\vspace{-.2cm}
	\item \textbf{Convolution.} Since geometry images inherit the geodesic neighborhood of the 3D object surface~\cite{cross_atlas}, convolution on the geometry image is directly equivalent to feature learning on the manifold surface of the 3D object.
	\vspace{-.2cm}
	\item \textbf{Padding.} Geometry image is analogous to cutting and unfolding the continuous 3D surface onto a regular 2D grid (the red lines in Figure~\ref{padding}). The boundaries of the geometry image are actually seamlessly connected with one another. If we rotate the geometry image 180 degrees around its central point, the rotated images can be seamlessly connected with the original geometry image~\cite{spherical_geometryimage,DL_geometryimage}. Therefore, to maintain the boundary continuity of the geometry image, the padding structure as shown in Figure~\ref{padding} is applied in this work. 
	\vspace{-.2cm}
	\item \textbf{Upsampling.} Benefiting from the geodesic neighborhood contained in the geometry image, upsampling of the geometry image is actually analogous to the upsampling of the 3D point cloud.
	\vspace{-.2cm}
\end{itemize}

Based on the above discussions, the architecture of our GIG can be as simple as the classical 2D image generation networks~\cite{DCGAN,Introvae} with our modified padding structure in Figure~\ref{padding}, while the output is a geometry image containing the coordinate values of the 3D point cloud, as shown in Figure~\ref{fig2}.

Notably, though our basic idea is motivated by the geometry image~\cite{gu2002geometry,DL_geometryimage}, the proposed GIG is different: 1) Instead of creating the geometry image by cutting and unfolding the 3D surface onto a 2D plane with complex algorithms, the proposed GIG achieves this with an end-to-end learning by dint of the training samples; 2) The output of our GIG is a more general 2D grid representation of the 3D object surface, and the original geometry image~\cite{spherical_geometryimage,DL_geometryimage} can actually be regarded as one particular solution of our generation model.

%-------------------------------------------------------------------------
\subsection{Optimization with Ad-VAE}
\label{ad-VAE}

The proposed encoder $\textit{E}$ and generator $\textit{G}$ are optimized using the ad-VAE which combines the adversarial learning of GAN with VAE. By adding additional adversarial losses to the encoder and generator, our ad-VAE not only leverages the fidelity of the generated point clouds, but also maintains the training stability of VAE.

Specifically, the objective function of our encoder and generator can be respectively formulated as follows:
\begin{align}
	\mathcal{L}_{\textit{E}} = \mathcal{L}_{rec} + \alpha \mathcal{L}_{prior} + \beta \mathcal{L}_{ad}^{\textit{E}}
\end{align}
\vspace{-.8cm}
\begin{align}
\mathcal{L}_{\textit{G}} = \mathcal{L}_{rec} + \beta \mathcal{L}_{ad}^{\textit{G}} 
\end{align}
where $\mathcal{L}_{rec}$ is the reconstruction loss between the input point cloud and the reconstructed result, the prior loss $\mathcal{L}_{prior} = {\rm KL}(q(z|\mathcal{P})||p(z))$ regularizes the input point clouds' latent representations to the prior distribution. $\mathcal{L}_{ad}^{\textit{E}}$ and $\mathcal{L}_{ad}^{\textit{G}}$ represent the adversarial loss which measures the distance between the prior distribution and the latent distributions of the reconstructed and generated fake point clouds. $\alpha$ and $\beta$ are the constant weights, and our ad-VAE will collapse to the standard VAE~\cite{VAE} with $\beta=0$.

Similar to GAN~\cite{GAN}, the proposed encoder and generator are also iteratively optimized. However, except for the point-wise reconstruction loss of VAE~\cite{VAE}, the introduced adversarial loss in our ad-VAE further encourages the generator to create plausible fake point clouds that have the same latent distribution as the input real point clouds, which can be seen as a feature level constraint to eliminate the blurring of the samples generated by VAE~\cite{VAE}. 

\textbf{Reconstruction loss.}
  The output of our generator $\textit{G}$ is a geometry image containing $n$ pixels, where each pixel corresponds to a 3D point with ${\rm x}$, ${\rm y}$, and ${\rm z}$ coordinate values. Thus, the generated geometry image can be directly regarded as a point cloud containing $n$ points. Denote the reconstructed point cloud as  $\mathcal{\tilde{P}}$, we use the Chamfer distance (Eq.~\ref{ChamferDistance}) to measure the difference between the input point cloud $\mathcal{P}$ and its reconstruction $\mathcal{\tilde{P}}$ as following
  \begin{equation}
  \mathcal{L}_{rec} =d_{CD} (\mathcal{P}, \mathcal{\tilde{P}})
  \end{equation}

\textbf{Adversarial loss.}
The adversarial loss is the key to embedding the adversarial learning into VAE. It encourages the encoder to not only map the real point clouds to the prior distribution, but also separate the latent distributions of the fake and real point clouds. The generator in turn is aimed at creating plausible fake point clouds which have the same latent distribution as the real point clouds.

Therefore, the adversarial loss of the generator and encoder can be formulated as:
\begin{align}
	\mathcal{L}_{ad}^{\textit{G}} = \frac{1}{2} \left( \!
			D_f \! \left(\! q(z|\mathcal{\tilde{P}}) || q(z|\mathcal{P}) \! \right) \! + \!
			D_f\! \left(\! q(z|\mathcal{\hat{P}})|| p(z) \!\right)
			\! \right) 
\end{align}
\vspace{-.5cm}
\begin{equation}
\mathcal{L}_{ad}^{\textit{E}} = -\mathcal{L}_{ad}^{\textit{G}}
\end{equation}
where $\mathcal{\hat{P}}$ is the generated point cloud and $D_f(q||p)$ represents the $f$-divergence~\cite{f_GAN} between distributions $q$ and $p$. 

Since the KL divergence has been used as the prior loss $\mathcal{L}_{prior}$, the intuitive choice for $D_f(q||p)$ should be the KL divergence. However, because of the unbounded gradient of the KL divergence, $\mathcal{L}_{ad}^{\textit{E}}$ also has no lower bound and is unstable to be minimized in practice. Therefore, the squared Hellinger distance~\cite{f_GAN} $H^2(q,p)\in[0,1]$ is finally applied in this work.

%-------------------------------------------------------------------------------
\section{Experiments}

\subsection{Evaluation Metrics}

We use the following metrics as in~\cite{l_GAN} to evaluate the generation capability of our method.

\iffalse
The \textit{Jensen-Shannon Divergence} (JSD) between the obtained empirical distributions $(P,Q)$:
\begin{equation}
	{\rm JSD}(P||Q) = \frac{1}{2} \left( {\rm KL}(P||M) + {\rm KL}(Q||M) \right)
\end{equation}
where $M=(P+Q)/2$.
\fi

The \textit{Jensen-Shannon Divergence} (JSD) measures the difference between the marginal distributions of the point clouds in the reference and generated sets. 

The \textit{Minimum Match Distance} (MMD) is the average distance between the generated point clouds and their nearest neighbors in the reference set. It measures the fidelity of the generated point clouds.

The \textit{Coverage} (COV) is the fraction of the point clouds in the reference set that are matched to the point clouds in the generated set. It measures the richness of the generated point clouds.

In this paper, the JSD is calculated by discretizing the point cloud space into $28^3$ voxels. The MMD and COV are implemented with both Chamfer and Earth-Mover distances, yielding four metrics: MMD-CD, MMD-EMD, COV-CD and COV-EMD. The higher COV means better performance, while the JSD and MMD are the opposite.

\begin{figure*}
	\setlength{\abovecaptionskip}{1cm} 
	\setlength{\belowcaptionskip}{-1.0cm} 
	\begin{center}
		\includegraphics[width=1.0\linewidth]{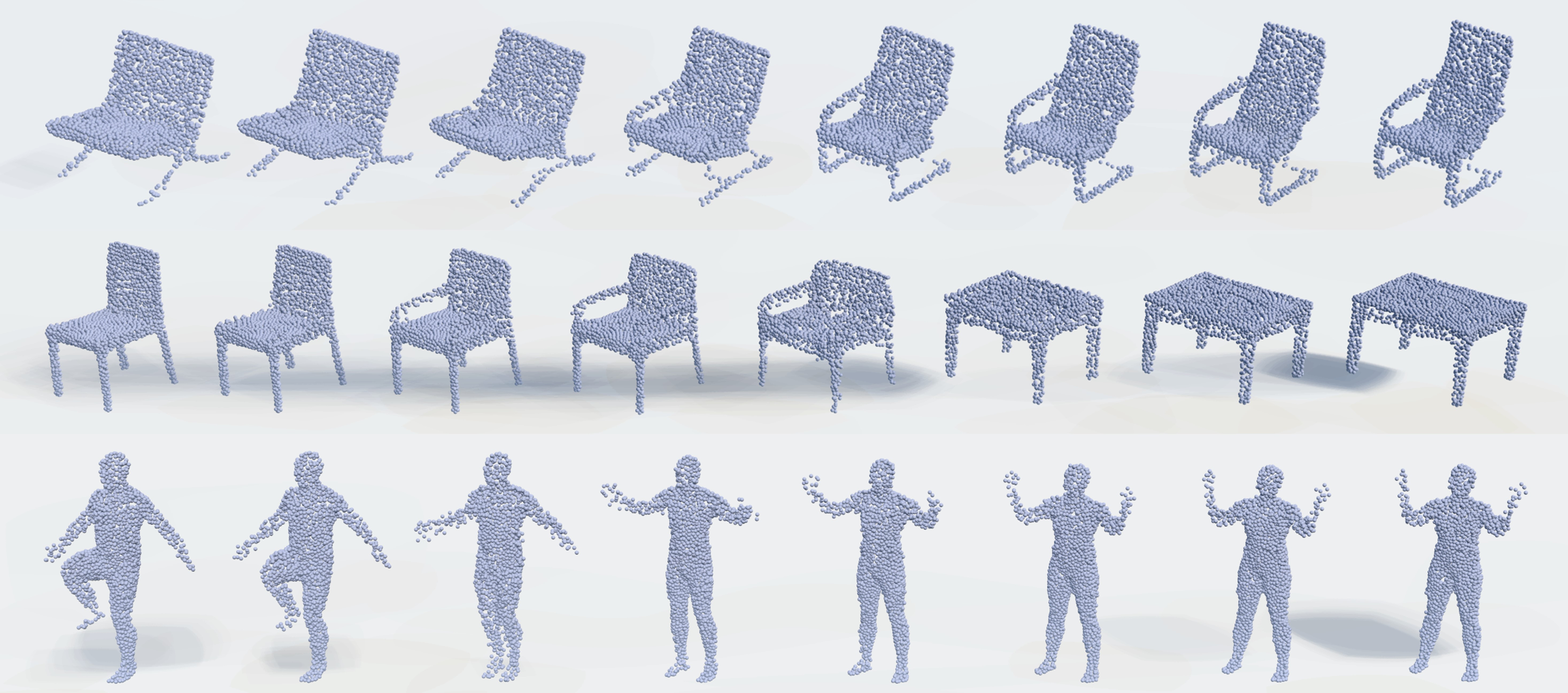}
	\end{center}
	\vspace{-0.3cm}
	\caption{Interpolation results in the latent space. \textbf{First row:} interpolation between two different chairs. \textbf{Second row:} translating a chair to a table.  \textbf{Third row:} converting a jumping man to a standing woman. All interpolation processes show smooth transitions. }
	\label{interpolation}
\end{figure*}
%-------------------------------------------------------------------------
\subsection{Generation Capabilities}

\newcommand{\tabincell}[2]{\begin{tabular}{@{}#1@{}}#2\end{tabular}}
\renewcommand{\arraystretch}{1.1} %control the height of the table
\begin{table}\renewcommand{\tabcolsep}{5pt} 
	\begin{center}
		\begin{tabular}{c|c c c c c}
			%\hline
			Method   	&JSD   	& \tabincell{c}{MMD- \\ CD}   	& \tabincell{c}{MMD- \\ EMD}     &\tabincell{c}{COV- \\CD}   	& \tabincell{c}{COV- \\EMD} \\ \hline  \hline
			r-WGAN    	&0.0450  &0.00154  	&0.0899  	  &63.31   		& 23.46 \\
			l-WGAN    	&0.0362  &0.00135  	&0.0636  	  &63.98   		& 26.27 \\
			Ours    	&\textbf{0.0341}  &\textbf{0.00133}  	&\textbf{0.0528}  	  &\textbf{65.30}   		& \textbf{39.79} \\			
			
		\end{tabular}
	\end{center}
	\vspace{-0.2cm}
	\caption{Generation results on the test dataset of the chair category. We compare the generation capability of our method with other methods including r-WGAN and l-WGAN~\cite{l_GAN} using FCN as the generator and decoder. 
	}
	\label{tab1}
\end{table}

\begin{table}\renewcommand{\tabcolsep}{.8pt}
	\begin{center}
		\begin{tabular}{c c c c c c c}
			\toprule
			\multirow{2}{*}{Category}&
			\multicolumn{2}{c}{JSD}&\multicolumn{2}{c}{Fidelity}&\multicolumn{2}{c}{Coverage} \\ 
			\cmidrule(lr){2-3} \cmidrule(lr){4-5} \cmidrule(lr){6-7}
			&l-WGAN &Ours &l-WGAN &Ours &l-WGAN &Ours \\

			\midrule
			Table   	&0.0788  &\textbf{0.0495}  	&0.0551  	  &\textbf{0.0484} 	& 14.53 &\textbf{34.13} \\
			Car    	&0.0318  &\textbf{0.0309}  	&0.0367  	  &\textbf{0.0295} 	&19.04 &\textbf{23.55} \\
			Airplane &0.0959  &\textbf{0.0926}  	&0.0365  	  &\textbf{0.0334} 	&20.05 &\textbf{20.92} \\
			Rifle    &0.0371  &\textbf{0.0340}  	&0.0448  	  &\textbf{0.0361} 	&18.11 &\textbf{25.47} \\
			
			\bottomrule
		\end{tabular}
	\end{center}
	\vspace{-0.2cm}
	\caption{JSD, Fidelity (MMD-EMD) and Coverage (COV-EMD)~\cite{l_GAN} results on the test dataset of table, car, airplane and rifle. 
	}
	\label{tab2}
\end{table}

We start by evaluating the generation capabilities of our method on both rigid and non-rigid 3D objects. After that, we extend it as a conditional ad-VAE to further explore its conditional generation capabilities for specific object categories.

\textbf{Rigid 3D object generation.} We use the 3D CAD models from ShapeNet~\cite{ShapeNet} to create our dataset for rigid object generation. Following~\cite{l_GAN}, we uniformly sample 2048 points from the mesh surface of each object to convert it to the corresponding point cloud. The same train/validation/test split (approximately 70\%-10\%-20\%) as that in ShapeNet~\cite{ShapeNet} is used for each category. 

We first evaluate our generation model on single object category. Five categories including chair, table, car, airplane, and rifle are considered in this experiment. For each category, a corresponding generator is trained and evaluated. To suppress the sampling bias of these measurements, each generator produces a set of point clouds that is $3\times$ the population of the test or validation set. This process is repeated three times and their averages are reported. 

In Table~\ref{tab1} we provide the generation results of our GIG on the chair category and compare them with r-WGAN and l-WGAN~\cite{l_GAN}. We can observe that the proposed GIG has achieved better performances across all metrics. In addition, quantitative evaluations on the other four categories, including table, car, airplane and rifle, also demonstrate the consistent results (Table~\ref{tab2}). Compared to l-WGAN using FCN for point cloud generation, the proposed GIG is more effective in capturing the geometric structure of the 3D point cloud, which leads GIG to be more effective in creating novel 3D point clouds with fine-grained structures. 

\textbf{Non-rigid 3D object generation.} We use the human models from the D-FAUST dataset~\cite{D_FAUST} to evaluate the generation capability of our GIG on non-rigid objects. The D-FAUST dataset contains approximately 40,000 meshes from 10 people. Each person have maximally 14 motions, and each motion is recorded by $\sim$300 meshes. For each person, we randomly select 100 meshes from his/her each motion, and spilt them into 70\%-10\%-20\% as train/validation/test dataset. For each mesh, we uniformly sample 2048 points to convert it into the point cloud.

Similar to the above experiments on rigid objects, we select the model that performs the best on the validation dataset and report its generation results on the test dataset in Table~\ref{tab3}. We can observe that our results on non-rigid objects have shown consistently better performances. In addition, compared to the point-wise loss of l-WGAN~\cite{l_GAN}, the adversarial loss of our ad-VAE further constraints the generated and real point clouds to have the same distribution in the latent space, which is helpful for creating plausible point clouds at semantic level.

\begin{table}\renewcommand{\tabcolsep}{5pt} 
	\begin{center}
		\begin{tabular}{c|c c c c c}
			%\hline
			Method   	&JSD   	& \tabincell{c}{MMD- \\ CD}   	& \tabincell{c}{MMD- \\ EMD}     &\tabincell{c}{COV- \\CD}   	& \tabincell{c}{COV- \\EMD} \\ \hline  \hline
			l-WGAN    	&0.1050  &0.0199  	&0.0937  	  &53.17  		&17.67 \\
			Ours    	&{\bf 0.0269}  &{\bf 0.0107}  	&{\bf 0.0513}  	 &{\bf 61.50}   & {\bf 35.16} \\			
			
		\end{tabular}
	\end{center}
	\vspace{-0.2cm}
	\caption{Generation results on non-rigid objects (human models from D-FAUST~\cite{D_FAUST}).
	}
	\label{tab3}
\end{table}

\begin{table}\renewcommand{\tabcolsep}{6pt} 
	\vspace{-.1cm} 
	\begin{center}
		\begin{tabular}{c|c c c}
			%\hline
			Method          	&JSD   	&Fidelity   	&Coverage  \\ \hline  \hline
			Ad-VAE          	&{\bf 0.0482}  &{\bf 0.0400}  	&28.77   \\
			Conditional ad-VAE  &0.0497  &0.0434  	&{\bf 31.50} \\			
			
		\end{tabular}
	\end{center}
	\vspace{-0.2cm}
	\caption{Generation results of the conditional ad-VAE. Here we report their average generation results over five categories including chair, table, car, airplane, and rifle. Their detailed results can be found in the Appendix.
	}
	\label{tab4}
\end{table}

\begin{figure*}
	\begin{center}
		\includegraphics[width=1.0\linewidth]{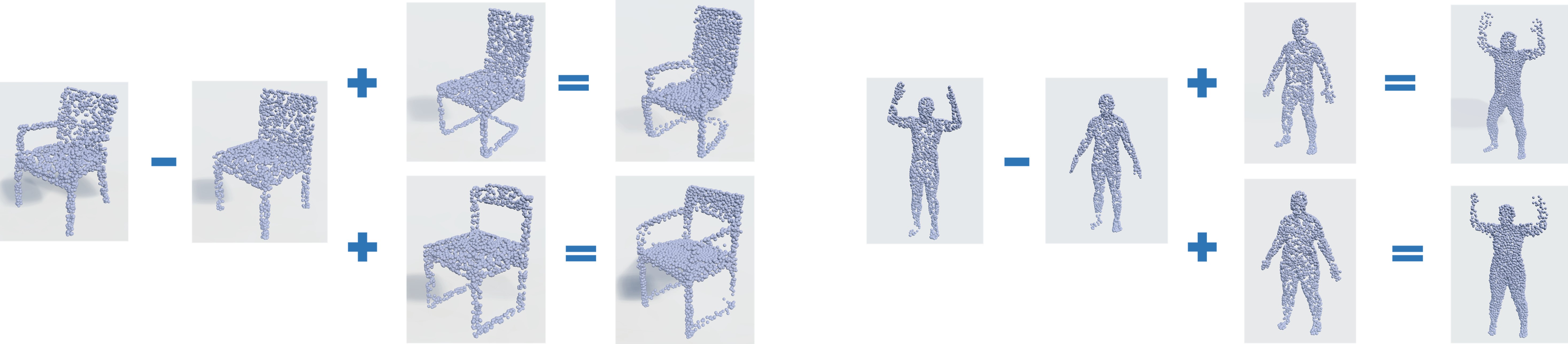}
	\end{center}
	\vspace{-0.3cm}
	\caption{Latent space arithmetic results. \textbf{Left:} adding arms to different chairs. \textbf{Right:} translating motion to different people.  }
	\label{arithmetic}
\end{figure*}
\textbf{Conditional 3D object generation.} In this section we further extend our ad-VAE (Section~\ref{method}) as a conditional ad-VAE for generating specific object categories. Similar to the conditional VAE~\cite{VAE_tutorial}, we encode the category label of each object as a one-hot vector and combine it with the latent representation as the conditional information to generate 3D point cloud of specific object category. In this experiment, five of the same categories as above are used to train a conditional generation model, and their average generation results are reported in Table~\ref{tab4}. 

Results show that the overall performance of our conditional ad-VAE is similar to the proposed ad-VAE (Section~\ref{method}) on single category. This implies that different object categories actually have similar structure properties which can be represented by a unified model. Meanwhile, because of the diversity and complexity of objects from multiple categories, the conditional ad-VAE tends to create more diverse objects (higher coverage) but with a slight fidelity discount.

%-------------------------------------------------------------------------
\subsection{Latent Representation}

To deeper understand the learned latent representations, experiments including interpolation and arithmetic between different objects are further conducted in this section. We can see that interpolation and arithmetic in the latent space can still show meaningful results.

\textbf{Latent space interpolation.}  In Figure~\ref{interpolation}, we provide the interpolation results between two different objects in the latent space. Previous works have demonstrated such interpolation on 2D images within the same category~\cite{DCGAN}. Here we show the interpolations for both rigid and non-rigid 3D objects, within and across object categories. Results show that the interpolation in the latent space involves smooth morph of the corresponding objects in point cloud space.

\textbf{Latent space arithmetic.} To further explore the semantic meaning of the latent representation, we also apply arithmetic on different objects in the latent space and observe how it affects the generated 3D objects. Specifically, we calculate the difference between two given objects (e.g., the arms of chair), in latent space, and add it to other objects. The edition results are provided in Figure~\ref{arithmetic}. We can observe that the arm can be equipped to other chairs by a simple addition operation on their latent representations, and the motion of a person can also be translated to other people.

%-------------------------------------------------------------------------
\subsection{3D Point Cloud Completion Capabilities}

In this section we further explore the effectiveness of our GIG on the 3D point cloud completion task~\cite{PCN}. The input is the observed partial point cloud and our goal is to generate the complete point cloud of its corresponding 3D object. 

\textbf{3D point cloud completion.} We choose eight object categories from ShapeNet~\cite{ShapeNet} to evaluate the completion capabilities of our GIG. They are the table, car, chair, airplane, sofa, rifle, lamp and watercraft. Our train/validation/test split is set to be the same as ShapeNet~\cite{ShapeNet}. For each 3D CAD model, we render its depth image from a random view and then back-project it into the 3D space to obtain its partial point cloud. The corresponding complete point cloud is uniformly sampled on the mesh surface. Similar to the previous generation experiments, 2048 points are sampled for each mesh.

\begin{figure}
	\begin{center}
		\includegraphics[width=1.0\linewidth]{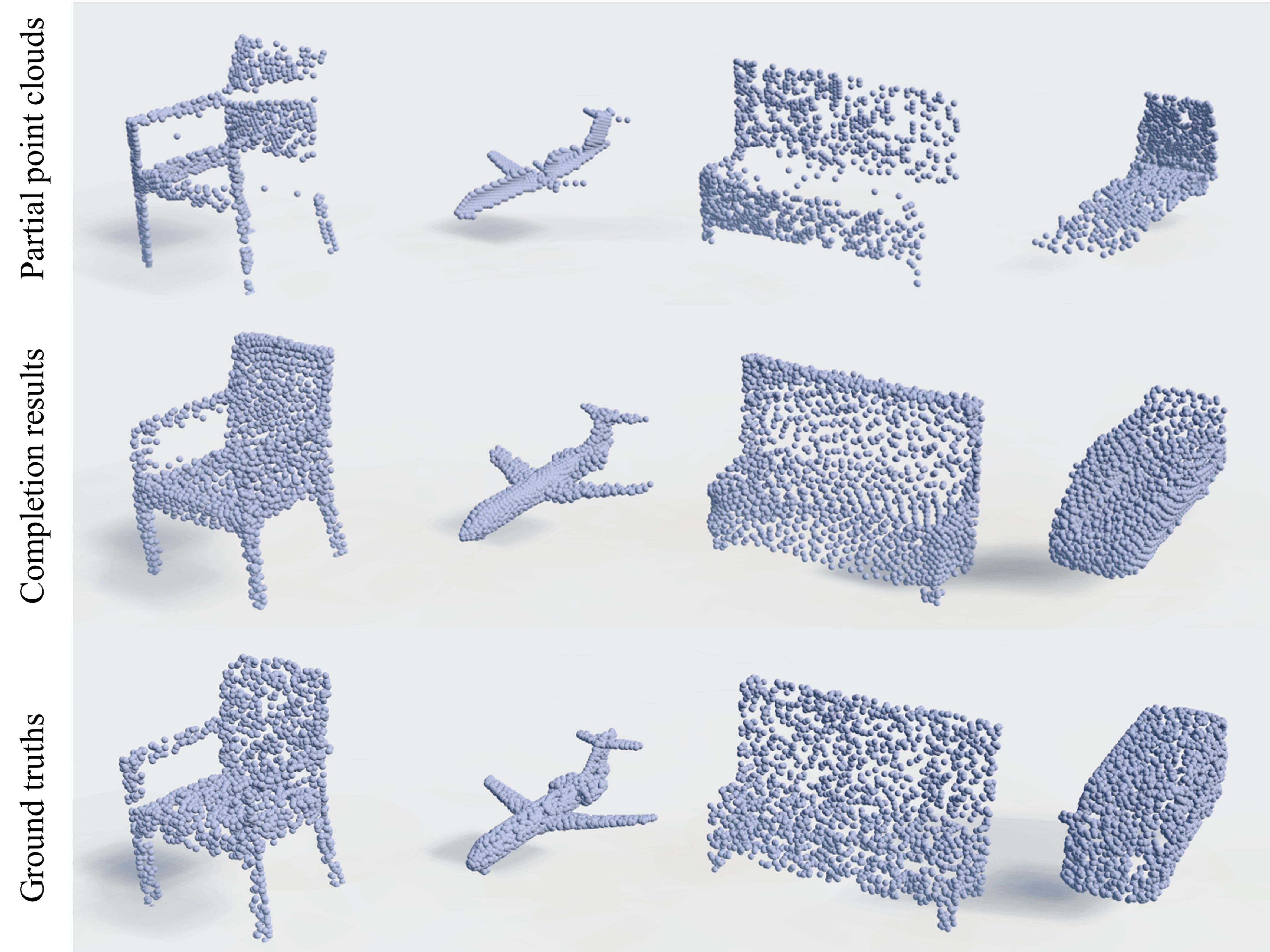}
	\end{center}
	%\vspace{-0.1cm}
	\caption{3D object completion results. Figures from top to bottom are the partial point clouds, the completion results and the ground truths. Notably, the table and airplane are from the test dataset, while the bench and bus are from the novel dataset which are not included in the training categories.  }
	\label{fig6}
\end{figure}

We compare our GIG with other state-of-the-art methods, including AtlasNet~\cite{AtlasNet}, FoldingNet~\cite{FoldingNet}, PCN~\cite{PCN}, and TopNet~\cite{TopNet}. For a fair comparison, the vanilla PointNet~\cite{PointNet} is used as the encoder for all experiments. Their loss function is calculated as the Chamfer distance between their outputs and the ground truths (i.e., the complete point clouds). All of them are trained with 300 epochs and a batch size of 32. Models that have the best performances on the validation dataset are chosen as the final models. We report their completion results in Table~\ref{3D_completion_results}, and show visual results in Figure~\ref{fig6}. We can seen that our GIG significantly outperforms the other comparison methods across all categories. It demonstrates that the proposed GIG not only can capture the geometric structure of the point cloud, but also has the capability to infer the unseen part of the 3D object.

\textbf{Completion for novel objects.} To further explore the completion capabilities of our GIG on unseen categories, we also create another dataset from other 3 categories including bench, bus, and pistol, called novel dataset. We randomly choose 200 CAD models from each category and generate their partial and complete point clouds as described above. Then the generators trained on the training dataset are applied on the novel dataset for 3D object completion. Their quantitative and qualitative results are provided in Table~\ref{3D_completion_results} and Figure~\ref{fig6} respectively. Results imply that the basic regulars such as symmetry and orthogonality learned by the generator are also applicable for novel 3D objects.

\begin{table*}\renewcommand{\tabcolsep}{3.5pt}
	\begin{center} \begin{threeparttable}
		\begin{tabular}{c c c c c c c c c c c c c c}
			\toprule
			\multirow{2}{*}{Method}&
			\multicolumn{9}{c}{Test dataset \tnote{*}}&\multicolumn{4}{c}{Novel dataset \tnote{*}} \\ 
			\cmidrule(lr){2-10} \cmidrule(lr){11-14} 
			&Table	&Car	&Chair	&Airplane	&Sofa	&Rifle	&Lamp	&Watercraft	&Avg	
			&Bench	&Bus	&Pistol	&Avg \\
			
			\midrule
			FCN~\cite{l_GAN}   &9.062	&5.296	&8.387	&4.319	&7.687	&3.255	&11.422	&5.692	&6.890	
			&8.568	&6.425	&7.570	&7.521 \\
			AtlasNet~\cite{AtlasNet} &8.508	&5.467	&8.837	&4.450	&7.649	&3.112	&9.663	&5.721	&6.682	
			&8.136	&6.641	&7.912	&7.563 \\
			FoldingNet~\cite{FoldingNet}	&7.971	&5.337	&8.137	&4.137	&7.319	&2.817	&7.840	&5.067	&6.078	
			&6.558	&4.885	&7.186	&6.210 \\
			TopNet~\cite{TopNet}	&8.005	&5.306	&8.099	&4.087	&7.269	&2.879	&8.946	&5.163	&6.219	
			&8.443	&4.681	&6.234	&6.453 \\
			PCN~\cite{PCN}	&7.647	&4.837	&7.320	&3.716	&6.385	&2.658	&8.480	&4.965	&5.751	
			&6.385	&5.020	&5.383	&5.596 \\
			Ours	&{\bf 6.767}	&{\bf 3.810}	&{\bf 5.417}	&{\bf 2.652}	&{\bf 5.126}	&{\bf 2.003}	&{\bf 7.670}	&{\bf 4.104}	&{\bf 4.694}	
			&{\bf 5.974}	&{\bf 3.855}	&{\bf 3.756}	&{\bf 4.528} \\

			\bottomrule
		\end{tabular}
	\begin{tablenotes}
		\footnotesize
		\item[*] The Chamfer distances are multiplied by $10^4$ for convenience sake, the smaller distance means better performance.
	\end{tablenotes}
	\end{threeparttable} \end{center}
	%\vspace{-0.2cm}
	\caption{3D point cloud completion results on ShapeNet. Notably, the test dataset contains the object categories which are same as the training dataset, while the novel dataset contains object categories which are unseen in the training dataset.
	}
	\label{3D_completion_results}
\end{table*}

%-------------------------------------------------------------------------
\subsection{Discussions on the Encoder and Generator}

To better understand the proposed encoder and generator (Section~\ref{network_architecture}), we further conduct several experiments to investigate the performance of the structure guided encoder and visualize the geometry images learned by our GIG.

\textbf{Effectiveness of the structure guided encoder.} The proposed structure guided encoder $\textit{E}$ needs to not only map the real point clouds to the prior distribution, but also distinguish the generated and reconstructed fake point clouds from the real point clouds. Both of them require the encoder to have powerful feature representation capabilities. To verify this, we further apply our encoder to the object classification task on ModelNet40~\cite{3d_ShapeNets} by replacing its last fully connected layer with Softmax classifier. Results in Table~\ref{performence_of_encoder} show that our encoder has achieved a +4.8\% accuracy by simply equipping the vanilla PointNet~\cite{PointNet} with the local geometric structure (Section~\ref{network_architecture}), which in turn verifies its feature representation capabilities. 

\begin{figure}
	\begin{center}
		\includegraphics[width=1.0\linewidth]{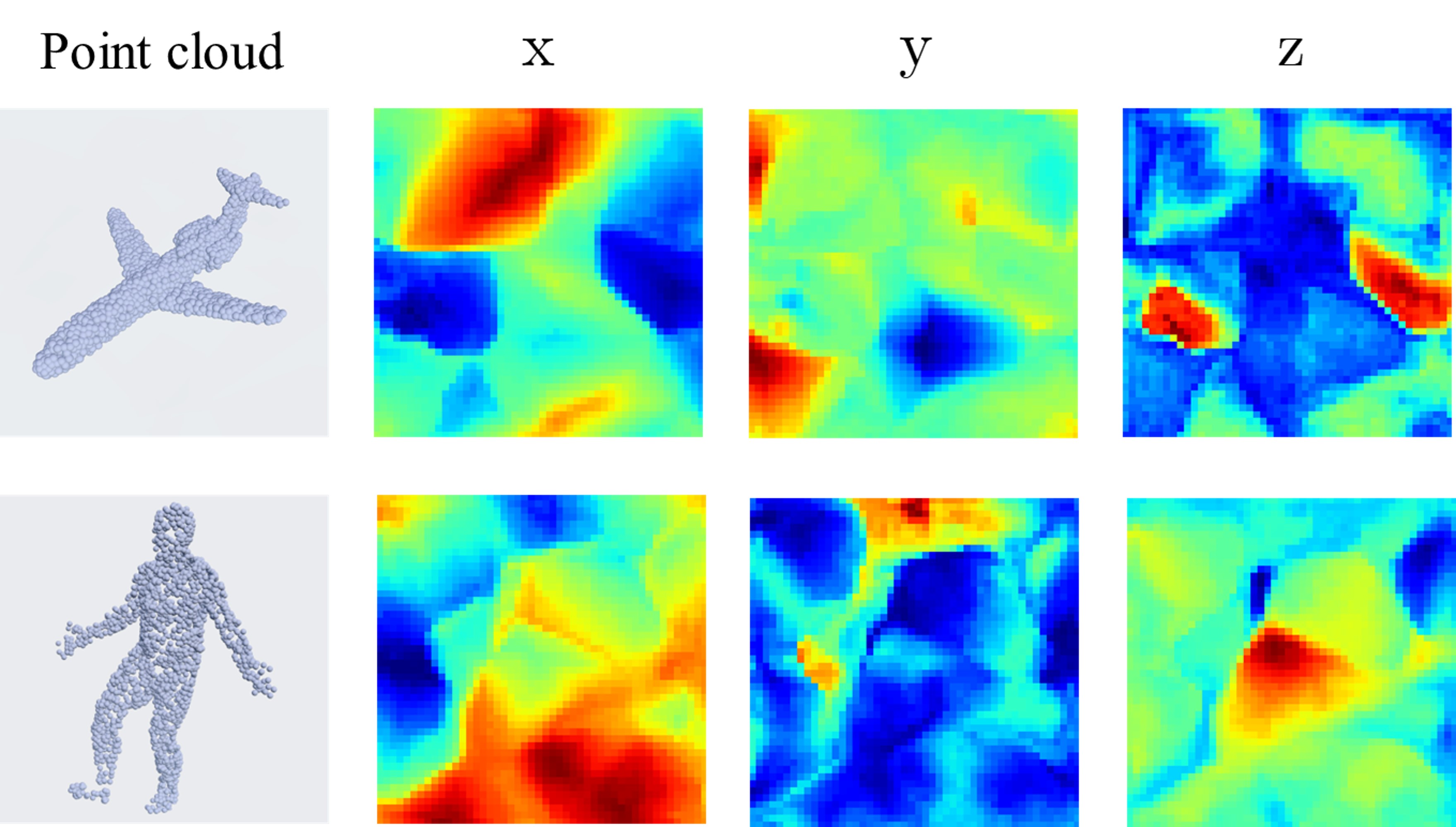}
	\end{center}
	\vspace{-0.1cm}
	\caption{Visualization of the generated point clouds and their corresponding geometry images with three channels. The output of our GIG is actually a more general 2D grid representation of the 3D point cloud, and the traditional geometry image~\cite{gu2002geometry,DL_geometryimage} can actually be regarded as one particular solution of our generation model. }
	\label{fig7}
\end{figure}

\textbf{Visualization of the learned geometry images.} Instead of creating the geometry image with traditional complex resampling algorithm~\cite{gu2002geometry,DL_geometryimage}, the proposed GIG actually achieves this with an end-to-end learning network. 
Thus the traditional  geometry image~\cite{spherical_geometryimage,DL_geometryimage} can be regarded as one particular solution of our generation model. To deeper understand what our GIG actually learned, we further provided more visualizations of the generated geometry images in Figure~\ref{fig7}. For visual convenience, each channel of the geometry image is rendered with a linear stretch.

\renewcommand{\arraystretch}{1.1} %control the height of the table
\begin{table}\renewcommand{\tabcolsep}{4pt} 
	\begin{center}
		\begin{tabular}{c|c c c}
			%\hline
			Method          	&PointNet (vanilla)~\cite{PointNet}   	&KCNet~\cite{KCNet}   	&Ours  \\ \hline  \hline
			Accuracy          	&87.1\%                 &91.0\%   &{\bf 91.9\%}  \\	
			
		\end{tabular}
	\end{center}
	\vspace{-0.1cm}
	\caption{
		Object classification results on the test dataset of ModelNet40~\cite{3d_ShapeNets}. %In comparison, our proposed encoder is more efficient to capture the local geometric structures which are helpful for the fine-grained 3D point cloud generation..
	}
	\label{performence_of_encoder}
\end{table} 

%-------------------------------------------------------------------------
\section{Conclusion}

We propose a novel geometry image based generator (GIG) for 3D point cloud generation. By converting the 3D point cloud generation problem to the regular 2D plane, the architecture of our GIG can be as simple as the 2D image generation networks. The experiments on diverse rigid and non-rigid 3D object datasets have demonstrated the effectiveness of our GIG for creating novel and plausible 3D point clouds.

%----------------------------------------
{\small
\bibliographystyle{ieee_fullname}
\bibliography{reference}
}

%----------------------------------------------------
%\include{Appendx}
\newpage
\appendix

%--------------------------------------------------------------------------------
\section*{Appendix}
%section A
\section{Overview}
In this document, we first provide additional details of the network architectures and training details of our proposed generation model in Section~\ref{secB}. More details of the conditional ad-VAE are provided in Section~\ref{secC}. Then the performance of the ad-VAE is further analyzed in Section~\ref{secD}, and some typical failure cases are provided in Section~\ref{secE}. Finally, more visualization results are presented in Section~\ref{secF}. 

%--------------------------------------------------------------------------------
%section B
\section{Network Architectures and Training Details}
\label{secB}

\textbf{Encoder.} Our structure guided encoder is similar to the architecture of the vanilla PointNet~\cite{PointNet}, but has a different input and output. For the input, we first calculate the point cloud's local geometric structure with $L=32$ dimensions, and then concatenate it with the ${\rm x}$, ${\rm y}$, ${\rm z}$ coordinates of the point cloud as the input of the encoder. Thus the input dimension of the encoder is actually 35. For the output, we modify the original classification network of PointNet to output the means and standard deviations of the latent representation with 128 dimensions.

\textbf{Generator.} Our geometry image based generator (GIG) is constructed by referring to the classical 2D image generator but with different padding structure (Section 3.1 of the main paper). The detailed parameters of our GIG are listed in Table~\ref{parameter_of_GIG}. Notably, as the output of our generator is a squared geometry image, we finally generate $46\times46$ points instead of 2048 points.

\textbf{Optimization of the ad-VAE.} In this work, we use the multivariate normal distribution $N(0,I)$ as the prior distribution $p(z)$. The outputs of our encoder are the latent representation's means and standard deviations $\mu$ and $\sigma$. Thus, our latent representation is calculated with reparameterization as $z=\mu + \epsilon \circ \sigma$, where $\epsilon \sim N(0,I)$ and $\circ$ is the point-wise product~\cite{VAE}.

\begin{table}[t] \renewcommand{\tabcolsep}{9pt} 
	%\vspace{.5cm}
	\begin{center}
		\begin{tabular}{c c c}
			\toprule
			Operations  	  &Parameters  	    &Output shape  \\ 
			\midrule \midrule
			
			Input             &-                  &1$\times$128  \\	
			FC 			  	  &128$\times$18432   &1$\times$18432  \\
			Reshape	          &-	              &512$\times$6$\times$6 \\
			Padding and Conv  &3$\times$3, 384	  &384$\times$6$\times$6 \\
			Padding and Conv  &3$\times$3, 384	  &384$\times$6$\times$6 \\
			Upsample	      &$\times$2	      &384$\times$12$\times$12 \\
			Padding and Conv  &3$\times$3, 256 	  &256$\times$12$\times$12 \\
			Padding and Conv  &3$\times$3, 256 	  &256$\times$12$\times$12 \\
			Upsample	      &$\times$2	      &256$\times$24$\times$24 \\
			Padding and Conv  &3$\times$3, 128 	  &128$\times$24$\times$24 \\
			Padding and Conv  &3$\times$3, 128 	  &128$\times$24$\times$24 \\
			Upsample	      &$\times$2	      &128$\times$48$\times$48 \\
			Padding and Conv  &3$\times$3, 128 	  &128$\times$48$\times$48 \\
			Padding and Conv  &3$\times$3, 128 	  &128$\times$48$\times$48 \\
			Conv	          &3$\times$3, 64	  &64$\times$46$\times$46 \\
			Conv and tanh	  &1$\times$1, 3	  &3$\times$46$\times$46 \\
			Reshape	          &-	              &3$\times$2116 \\
			
			\bottomrule
		\end{tabular}
	\end{center}
	\vspace{-0.1cm}
	\caption{
		Parameter details of the proposed GIG.
	}
	\label{parameter_of_GIG}
\end{table}

Similar to GAN, our encoder and generator are also iteratively optimized. Denote the parameters of the encoder and generator as $\Theta_{\textit{E}}$ and $\Theta_{\textit{G}}$ respectively, the training process of our ad-VAE is provided in Algorithm \ref{train_ad_VAE}. $q(z|\mathcal{P})$, $q(z|\mathcal{\tilde{P}})$ and  $q(z|\mathcal{\hat{P}})$ are the corresponding latent distributions of the real, reconstructed and generated point clouds respectively.

Notably, since the distance between $q(z|\mathcal{\hat{P}})$ and $q(z|\mathcal{P})$ can not be directly calculated in each training batch, we instead turning to minimize the distance between $q(z|\mathcal{\hat{P}})$ and the prior normal distribution $N(0,I)$, which is also the destination of $q(z|\mathcal{P})$. Thus the upper bound of the distance between $q(z|\mathcal{\hat{P}})$ and $q(z|\mathcal{P})$ is equally minimized.

\textbf{Training Details.} Our generation models are trained using the Adam optimizer with an initial learning rate of 0.0001, $\alpha=1$ and $\beta=0.1$. For the generation task on the ShapeNet~\cite{ShapeNet} and D-FAUST~\cite{D_FAUST} datasets, we train the model with 1200 epochs and batch size 32. To accelerate the convergence, the models for generation task are pretrained with 300 epochs using the VAE loss~\cite{VAE} (i.e., $\beta=0$). In addition, the coordinate values of the human models of the D-FAUST dataset are divided by 1.5 to make sure they are lying in a unit sphere. For the 3D point cloud completion task (Section 4.4), we train the model with 300 epochs and batch size 32. 

\begin{algorithm}[t]
	\caption{Training ad-VAE}
	\label{train_ad_VAE}
	$\Theta_{\textit{E}}$, $\Theta_{\textit{G}} \gets$ initialize network parameters \\
	{\bf repeat}
	
	\quad $\mathcal{P} \gets$ random mini-batch from dataset \\
	\vspace{-.0cm}
	\quad $\mu, \sigma \gets \textit{E} (\mathcal{P})$ \\
	\vspace{-.0cm}
	\quad $z \gets \mu + \epsilon \circ \sigma$ \{Reparameterization.\} \\
	\vspace{-.0cm}
	\quad $z_p \gets$ samples from prior distribution $N(0,I)$ \\
	\vspace{-.0cm}
	\quad $\mathcal{\tilde{P}}, \mathcal{\hat{P}} \gets \textit{G}(z), \textit{G}(z_p)$ \\
	\vspace{-.0cm}
	\quad $\mathcal{L}_{rec} \gets d_{CD}(\mathcal{P}, \mathcal{\tilde{P}})$ \\
	\vspace{-.0cm}
	\quad $\mathcal{L}_{prior} \gets {\rm KL}\left(q(z|\mathcal{P}) || N(0,1) \right)$ \\
	\vspace{-.0cm}
	\quad $\mathcal{L}_{ad}^{\textit{E}} \! \gets \! -\frac{1}{2} \! \left( \!
	H^2 \! \left(\! q(z|\mathcal{\tilde{P}}), q(z|\mathcal{P}) \! \right) \! + \!
	H^2\! \left(\! q(z|\mathcal{\hat{P}}), N(0,1) \!\right)
	\! \right) $ \\
	\vspace{-.0cm}
	\quad $\mathcal{L}_{ad}^{\textit{G}} \! \gets \! \frac{1}{2} \! \left( \!
	H^2 \! \left(\! q(z|\mathcal{\tilde{P}}), q(z|\mathcal{P}) \! \right) \! + \!
	H^2\! \left(\! q(z|\mathcal{\hat{P}}), N(0,1) \!\right)
	\! \right) $ \\
	\vspace{-.0cm}
	\quad $\Theta_{\textit{E}} \stackrel{+}\gets -\nabla_{\Theta_{\textit{E}}} (\mathcal{L}_{rec} + \alpha \mathcal{L}_{prior} + \beta \mathcal{L}_{ad}^{\textit{E}} )$ \\
	\vspace{-.0cm} 
	\quad $\Theta_{\textit{G}} \stackrel{+}\gets -\nabla_{\Theta_{\textit{G}}} (\mathcal{L}_{rec} + \beta \mathcal{L}_{ad}^{\textit{G}} )$ \\
	{\bf end}
	
\end{algorithm}

\begin{figure*}
	\begin{center}
		\includegraphics[width=1.0\linewidth]{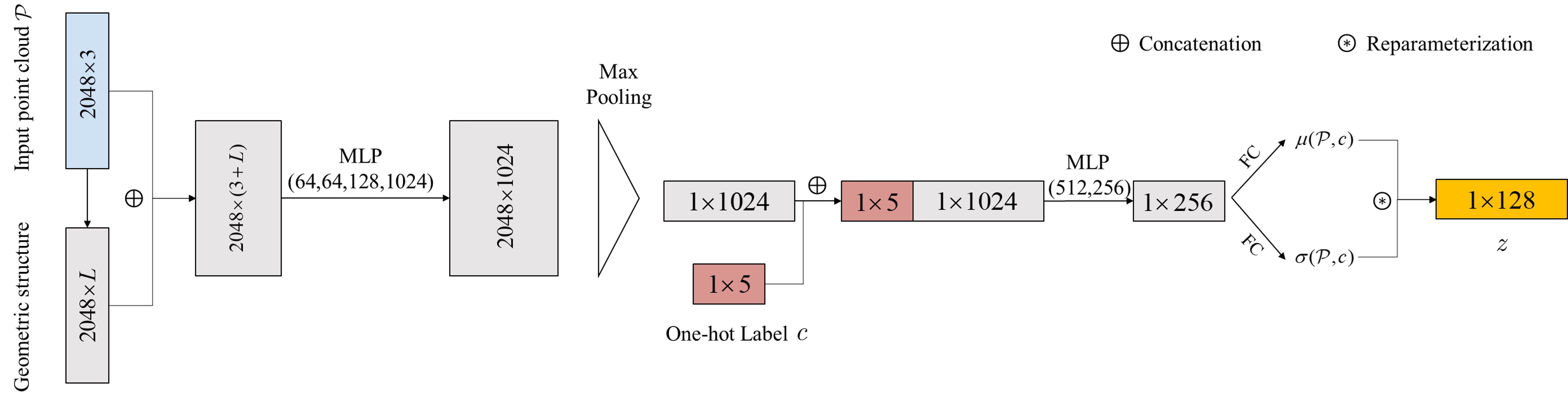}
	\end{center}
	\vspace{-0.3cm}
	\caption{Encoder architecture of the conditional ad-VAE (Section 4.2). The one-hot label is a vector with 5 channels, corresponding to the 5 object categories including chair, table, car, airplane and rifle.  }
	\label{c_ad_VAE}
\end{figure*}

%--------------------------------------------------------------------------------
%section C
\section{More Details of the Conditional Ad-VAE}
\label{secC}

\begin{figure}
	\begin{center}
		\includegraphics[width=1.0\linewidth]{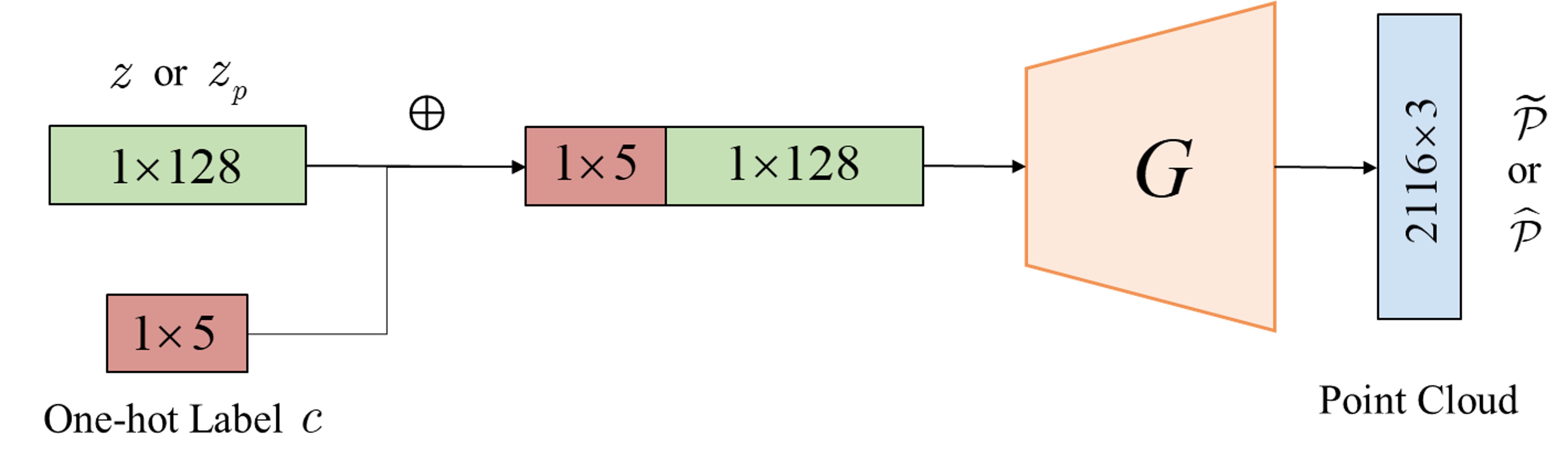}
	\end{center}
	\vspace{-0.3cm}
	\caption{Generator architecture of the conditional ad-VAE.  }
	\label{c_GIG}
\end{figure}

The architecture of our conditional ad-VAE is modified from the proposed ad-VAE by referring to the conditional VAE~\cite{VAE_tutorial}. However, as the conditional information here is the one-hot label which can not be directly combined with the point cloud and inputted to the encoder, we therefore embed the one-hot label into the encoder by concatenating it with the global feature of the point cloud, as shown in Figure~\ref{c_ad_VAE}. Correspondingly, the one-hot label is also concatenated with the latent representation as the conditional information to generate objects of specific category (Figure~\ref{c_GIG}). 

In Table~\ref{detailed_c_ad_VAE} we provide more detailed generation results of the conditional ad-VAE (Section 4.2) and compare them with the results of our ad-VAE on single category. We can see that these two experiments have achieved similar quantitative results, which reveals that different objects actually have common basic characteristics such as symmetry and orthogonality.

%--------------------------------------------------------------------------------
%section D
\section{Effectiveness of the Ad-VAE}
\label{secD}

\begin{figure*}
	\begin{center}
		\includegraphics[width=.96\linewidth]{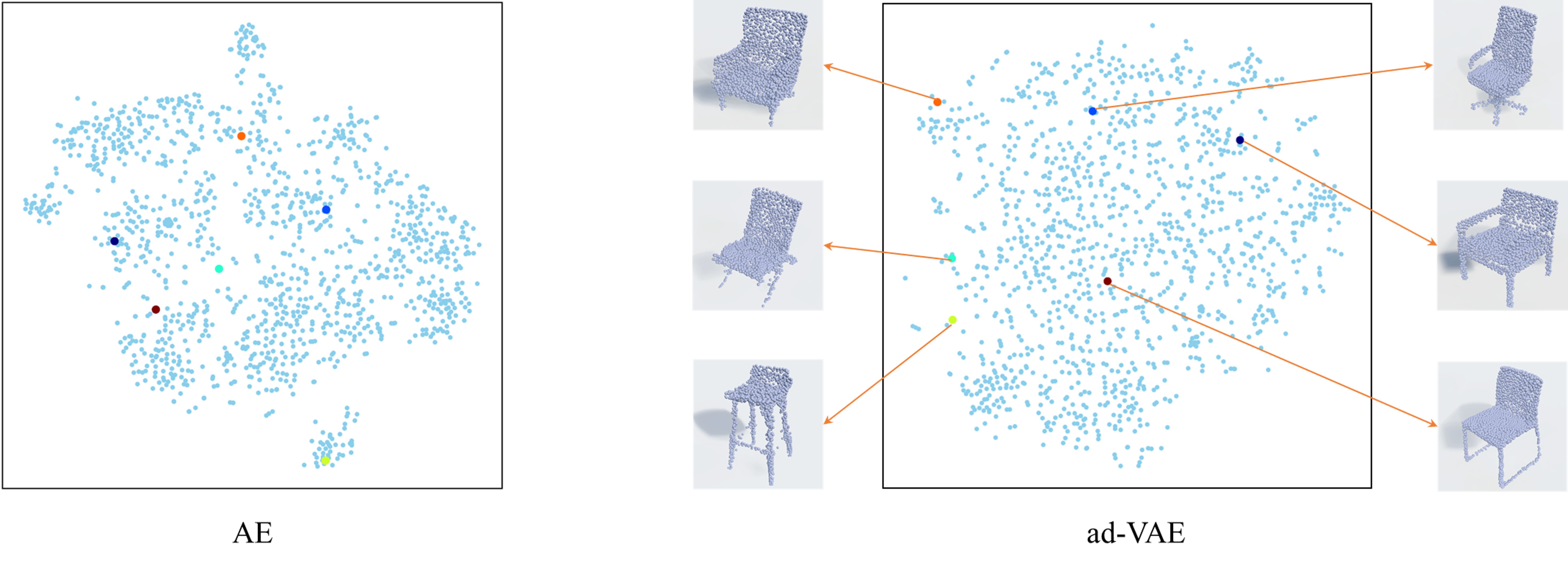}
	\end{center}
	\vspace{-0.3cm}
	\caption{Visualization of the latent representations using t-SNE~\cite{t_SNE} on the test dataset of the chair category. Left and right are the latent space trained with AE~\cite{l_GAN} and the proposed ad-VAE, respectively.   }
	\label{latent_ad_VAE}
\end{figure*}

In this section we further explore the effectiveness of the proposed ad-VAE by comparing it with the autoencoder (AE)~\cite{l_GAN}. In Figure~\ref{latent_ad_VAE} we show the distribution of the latent representations encoded by ad-VAE and AE using t-SNE~\cite{t_SNE}. We can observe the distribution of the proposed ad-VAE is more compact and smoother than AE, which is helpful for obtaining meaningful latent interpolation results.

%--------------------------------------------------------------------------------
%section E
\section{Failure Cases}
\label{secE}

\begin{figure*}
	\begin{center}
		\includegraphics[width=0.8\linewidth]{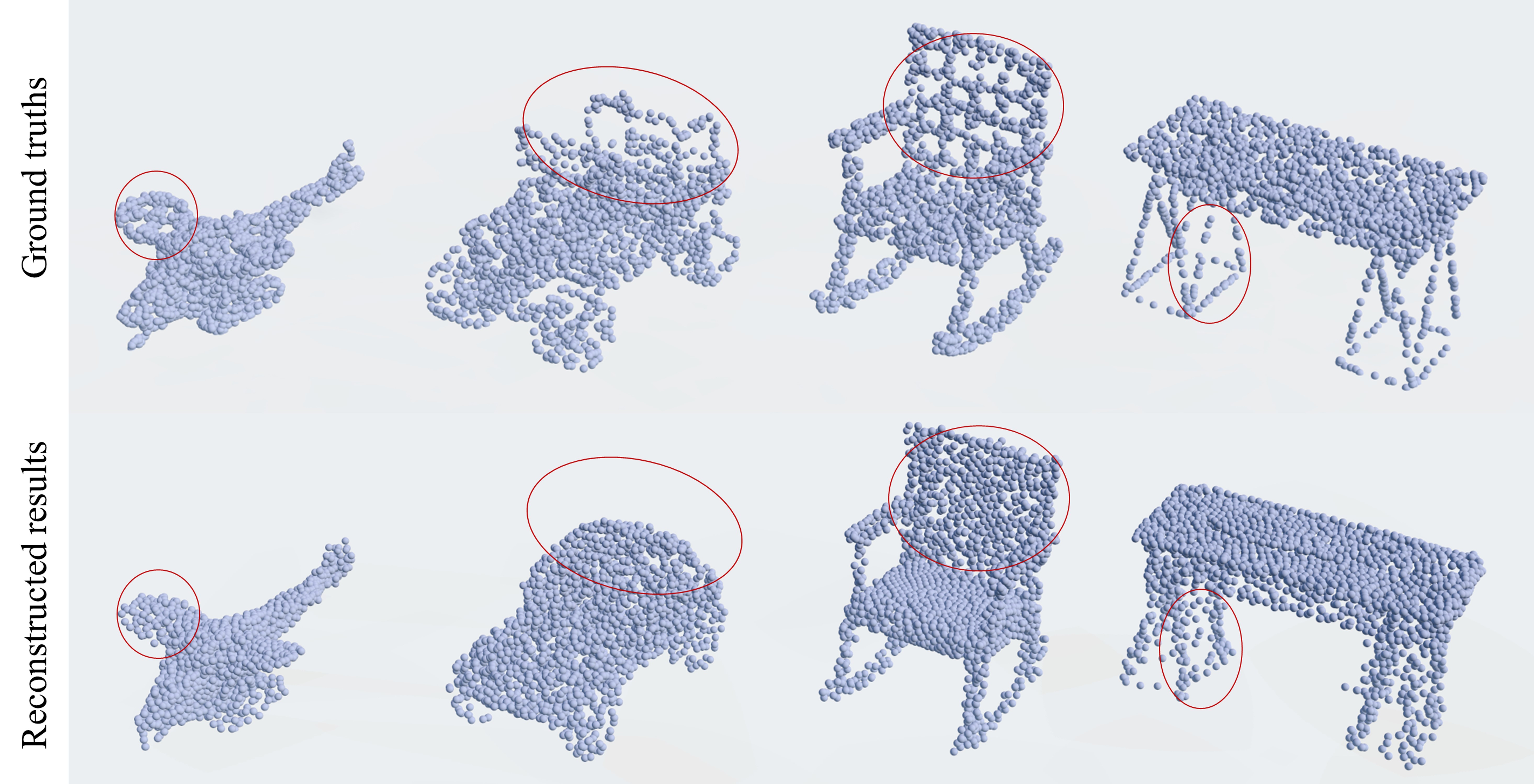}
	\end{center}
	\vspace{-0.3cm}
	\caption{Failure cases. The first row are the ground truths and the second row are the reconstructed results.  }
	\label{failure_cases}
\end{figure*}

In Figure~\ref{failure_cases} we show some typical failure cases. Specifically, objects with rich thin structures or rare shapes are difficult to reconstruct or generate.

\begin{table}\renewcommand{\tabcolsep}{3pt}
	\begin{center}
		\begin{tabular}{c c c c c c c}
			\toprule
			\multirow{2}{*}{Category}&
			\multicolumn{2}{c}{JSD}&\multicolumn{2}{c}{Fidelity}&\multicolumn{2}{c}{Coverage} \\ 
			\cmidrule(lr){2-3} \cmidrule(lr){4-5} \cmidrule(lr){6-7}
			&A &B &A &B &A &B \\
			
			\midrule
			Chair    &{\bf0.0341}	&0.0373	  &{\bf0.0528}	&0.0568	 &{\bf39.79}	 &37.36 \\
			Table    &{\bf0.0495}	&0.0496	  &{\bf0.0484}	&0.0540	 &34.13	 &{\bf35.08} \\
			Car    	 &{\bf0.0309}	&0.0311	  &{\bf0.0295}	&0.0301	 &23.55	 &{\bf29.62} \\
			Airplane &{\bf0.0926}	&0.0920	  &{\bf0.0334}	&0.0361	 &20.92	 &{\bf24.51} \\
			Rifle    &{\bf0.0340}	&0.0383	  &{\bf0.0361}	&0.0399	 &25.47	 &{\bf30.95} \\
			Avg      &{\bf0.0482}	&0.0497	  &{\bf0.0400}	&0.0434	 &28.77	 &{\bf31.50} \\
			
			\bottomrule
		\end{tabular}
	\end{center}
	\vspace{-0.2cm}
	\caption{
		Detailed generation results of the conditional ad-VAE. A and B are the ad-VAE and the corresponding conditional ad-VAE respectively. 
	}
	\label{detailed_c_ad_VAE}
\end{table} 

%--------------------------------------------------------------------------------
%section F
\section{More Visualizations}
\label{secF}

\begin{figure*}
	\begin{center}
		\includegraphics[width=1.0\linewidth]{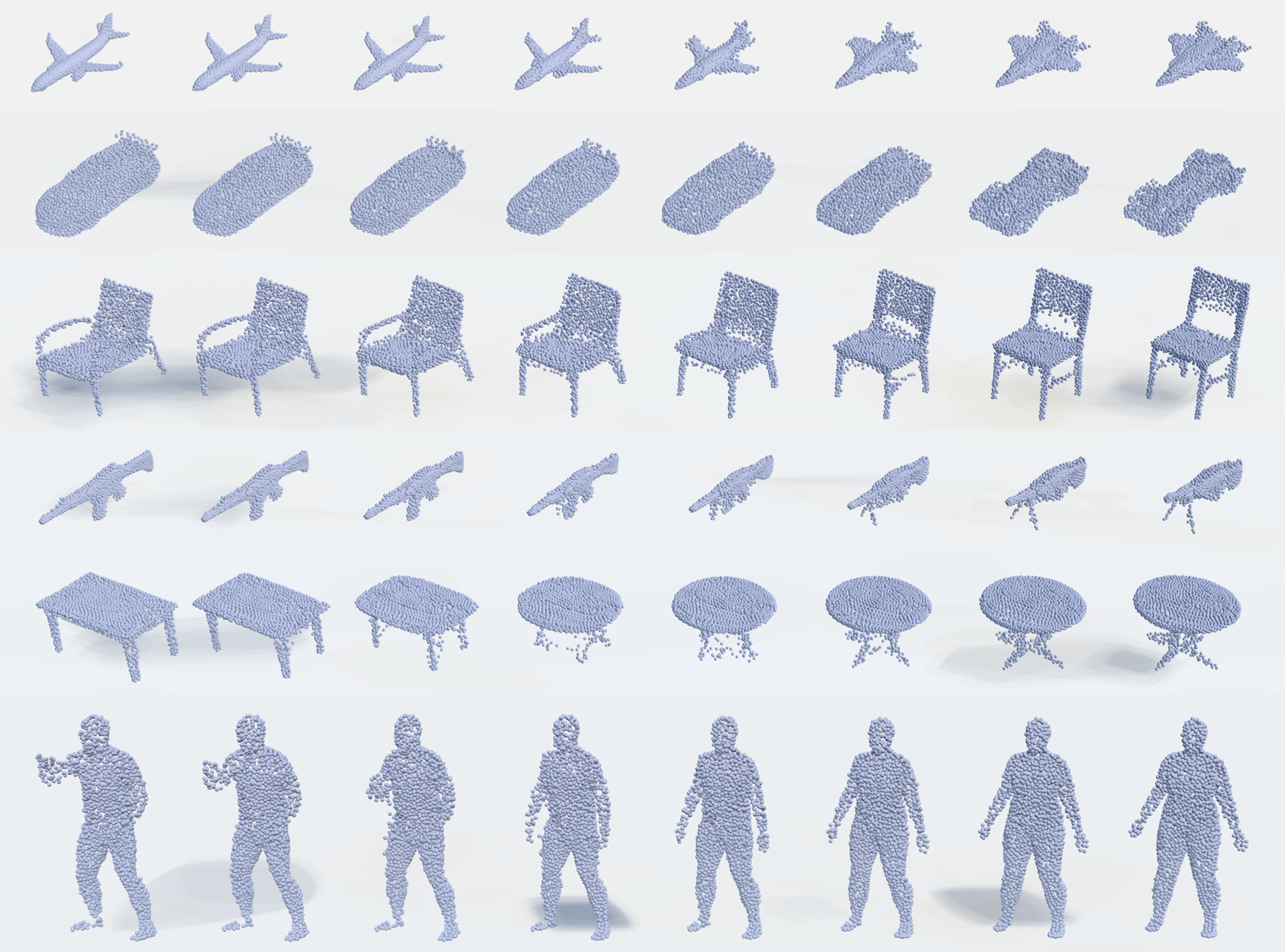}
	\end{center}
	\vspace{-0.3cm}
	\caption{More interpolation results on different objects. Figures from top to bottom are the airplane, car, chair, rifle, table and human model respectively.  }
	\label{more_interpolations}
\end{figure*}

\begin{figure*}
	\begin{center}
		\includegraphics[width=1.0\linewidth]{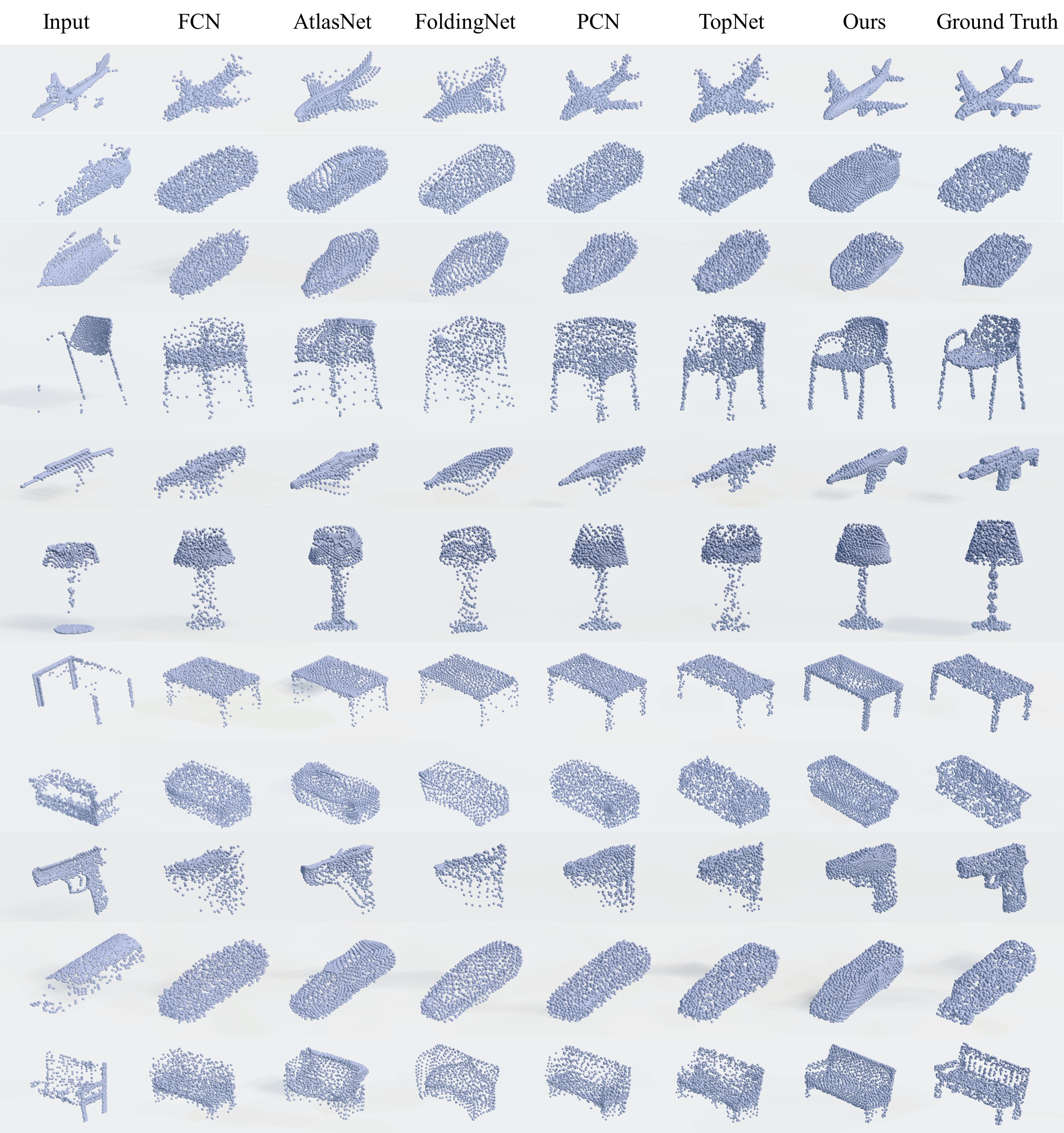}
	\end{center}
	\vspace{-0.3cm}
	\caption{3D completion results. The objects of the first eight rows are from the test dataset which contains the same categories as the training dataset, while the last three rows are from the novel dataset which contains categories unseen during training.  }
	\label{more_3D_completions}
\end{figure*}

In this section we provided more visualizations about the interpolation results in the latent space (Figure~\ref{more_interpolations}) and the 3D completion results (Figure~\ref{more_3D_completions}).  

\end{document}